\documentclass[11pt]{article}
\usepackage[margin=1in]{geometry}
\usepackage{amsmath,amsfonts,amssymb}
\usepackage{graphicx}
\usepackage{hyperref}
\usepackage{subcaption}
\usepackage{wrapfig}
\usepackage{amsmath,amssymb}
\usepackage{booktabs}
\usepackage{adjustbox}
\usepackage{makecell}
\usepackage{multirow}
\usepackage{caption}
\usepackage{xcolor}
\usepackage[table]{xcolor}

\usepackage{pgfplotstable}
\usepackage{booktabs}

\definecolor{bestRed}{HTML}{F8CECC}   
\definecolor{worstYellow}{HTML}{FFFFBF} 

\newcommand{\cc}[1]{\cellcolor{bestRed!#1!worstYellow}}

\newcommand{\instA}{\thanks{Institute of Robotics and Machine Intelligence, Poznan University of Technology}}

\title{A Comparative Evaluation of Geometric Accuracy in NeRF and Gaussian Splatting}
\author{Miko{\l}aj Zieli\'nski\instA\footnotemark[1] \and Eryk Vykysal\'y\footnotemark[1] \and Bart{\l}omiej Biesiada\footnotemark[1] \and Jan Baturo\footnotemark[1] \and Mateusz Capa{\l}a\footnotemark[1] \and Dominik Belter\footnotemark[1]}

\date{}

\begin{document}
\maketitle

\begin{abstract}
    Recent advances in neural rendering have introduced numerous 3D scene representations. Although standard computer vision metrics evaluate the visual quality of generated images, they often overlook the fidelity of surface geometry. This limitation is particularly critical in robotics, where accurate geometry is essential for tasks such as grasping and object manipulation. In this paper, we present an evaluation pipeline for neural rendering methods that focuses on geometric accuracy, along with a benchmark comprising 19 diverse scenes. Our approach enables a systematic assessment of reconstruction methods in terms of surface and shape fidelity, complementing traditional visual metrics.
\end{abstract}

\smallskip

\bigskip
\small\noindent\textit{This work has been accepted for publication in the proceedings of PP-RAI 2026 (Springer Lecture Notes in Networks and Systems). The final authenticated version will be available via Springer.}

\section{Introduction}

Accurate 3D scene reconstruction is a cornerstone of modern computer vision and robotics. Over the past decade, advances in neural rendering and implicit scene representations have enabled high-quality visual reconstructions of complex environments. Methods such as Neural Radiance Fields (NeRF)~\cite{Mildenhall2022} and their variants~\cite{instant-ngp} can produce photo-realistic renderings from sparse input views, often surpassing traditional multi-view reconstruction techniques in visual fidelity.
Despite these achievements, the evaluation of reconstructed scenes has focused mainly on image-based metrics~\cite{Kerbl2023gaussian_splatting,Mildenhall2022}, such as peak signal-to-noise ratio (PSNR) or structural similarity (SSIM), which quantify the similarity between rendered images and ground truth photographs. Although these metrics effectively measure appearance quality, they provide limited insight into the geometric fidelity of reconstructed surfaces—a factor that is crucial for downstream applications, including robotic manipulation, navigation, and physical interaction \cite{wang2025nerfsroboticssurvey,byravan2023nerf2real,rashid2023lerftogo,tseng2022clanerf,ichnowski2022dexnerf}.

Neural Radiance Fields provide a continuous volumetric representation of space~\cite{Mildenhall2022}, which constitutes one of their key advantages. This formulation enables querying the scene at arbitrary spatial locations, allowing flexible sampling in regions of interest. Such properties are particularly attractive in robotics, where geometry can be sampled densely around potential grasping regions or contact points \cite{wang2025nerfsroboticssurvey,ichnowski2022dexnerf}. However, despite these benefits, NeRF-based methods are computationally expensive to train and infer (render the scene).

To address these limitations, Gaussian Splatting~\cite{Mildenhall2022,Kerbl2023gaussian_splatting} was introduced as an alternative explicit scene representation. Instead of encoding the scene in a continuous neural field, it models geometry using a collection of optimized 3D Gaussian primitives whose positions, orientations, shapes, and opacities are learned during training. This explicit formulation offers two practical advantages: faster optimization and rendering and direct access to geometric primitives. Notably, the centers (means) of the optimized Gaussians form a dense point cloud approximating the object surface. This representation can be easily integrated with classical grasp planning and geometric processing pipelines~\cite{Zheng2024gaussiangrasper}.

Although there is now a wide range of scene representation methods, little attention has been paid to their ability to faithfully reconstruct object geometry. In particular, to the best of our knowledge, there is no widely adopted benchmark specifically designed to evaluate the quality of geometric reconstruction in a robotics relevant setting. To address this gap, we propose a benchmark comprising 19 diverse scenes containing YCB objects~\cite{Calli2017ycb}. Scenes are registered in a real-world metric coordinate system, enabling direct geometric comparisons in physical units. Using this benchmark, we evaluated several widely used scene representation models both qualitatively and, more importantly, quantitatively with respect to surface and shape fidelity. Our study aims to answer a fundamental question: which modern neural scene representations most accurately capture real-world geometry and are therefore best suited for robotic applications.

\section{Literature Review}

\subsection{Implicit NeRF-based Methods}

Implicit Neural Radiance Field (NeRF) approaches have become a dominant paradigm for novel view synthesis. \cite{Liao2025} provides a comprehensive study of NeRF and its variants. For camera pose estimation, COLMAP~\cite{schoenberger2016sfm,pan2024glomap} is widely used as a standard Structure-from-Motion (SfM) pipeline, providing reliable poses for multi-view reconstruction. Practical implementations, such as Nerfacto provided by Nerfstudio~\cite{nerfstudio_2023}, integrate many state-of-the-art techniques, including advanced density field sampling, camera pose refinement, and piecewise sampling strategies. Instant-NGP~\cite{instant-ngp} introduces a compact neural network augmented with a multiresolution hash table of trainable feature vectors, resulting in a simple and fast architecture that is parallelized on modern GPUs.


\subsection{Gaussian Splatting}

Gaussian Splatting~\cite{Kerbl2023gaussian_splatting} has emerged as a complementary representation for volumetric radiance fields. Kerbl et al.~\cite{Kerbl2023gaussian_splatting} represent scenes using 3D Gaussians that preserve the desirable properties of continuous volumetric radiance fields while avoiding unnecessary computation in empty regions. Their approach combines interleaved optimization and density control of 3D Gaussians, including anisotropic covariance optimization, alongside a fast visibility-aware rendering algorithm that supports anisotropic splatting for real-time rendering. \cite{Fei2025} provides a survey of existing Gaussian Splatting methods.


Liu et al.~\cite{Liu2025DBS} extend Gaussian Splatting by replacing Gaussian kernels with deformable Beta Kernels, which provide bounded support and adaptive frequency control for capturing fine geometric details more efficiently. Additionally, the Beta Kernel is applied to color encoding, improving the representation of both diffuse and specular components. Addressing limitations of standard Gaussians in modeling hard edges and flat surfaces, Held et al.~\cite{Held2025} introduce 3D Convex Splatting (3DCS), which uses smooth 3D convex primitives to construct geometrically meaningful radiance fields from multi-view images. Compared to Gaussians, convex shapes offer greater flexibility, enabling accurate representation of sharp edges and dense volumes with fewer primitives.

\subsection{Alternative Scene Representation and Rendering Methods}

Beyond Gaussian-based representations, several alternative methods exist. Triangle-based splatting~\cite{triangle_splatting} uses differentiable triangles for geometric efficiency and adaptive density, achieving higher visual fidelity, faster convergence, and greater rendering throughput than 2D/3D Gaussian splats. Radiance meshes~\cite{mai2026radiancemeshes} model scenes with constant-density Delaunay tetrahedra, natively supported by graphics hardware, enabling exact volume rendering via rasterization or ray tracing. Sparse voxel methods~\cite{sparse_voxels} render radiance fields with adaptive voxels without neural networks or Gaussian primitives. 2D Gaussian approaches~\cite{2d_gaussian} collapse the 3D volume into oriented planar disks, preserving view-consistent geometry across viewpoints.


NeRF- and Gaussian Splatting-based methods are widely used in robotics~\cite{wang2025nerfsroboticssurvey} for applications such as navigation~\cite{byravan2023nerf2real} and robotic manipulation~\cite{rashid2023lerftogo,tseng2022clanerf,ichnowski2022dexnerf}. In these applications, scene models are often evaluated primarily in terms of rendering accuracy or speed. However, in robotics, accurate 3D scene geometry is critical and is frequently underestimated in research. Furthermore, many studies claim that specific methods are more accurate than others considering 3D models without providing supporting evidence. In this paper, we present a pipeline, a dataset, and an experimental evaluation that rigorously assesses the 3D accuracy of the most popular scene rendering methods.

\section{Methodology}

    \subsection{Scene Capturing}
        To ensure accurate and repeatable data acquisition, we constructed a controlled capture setup using a \textit{UR5} robotic manipulator. An \textit{Orbbec Bolt} RGB-D camera was rigidly mounted on a handheld stick placed on the table. The stick was designed to be grasped by the robot, enabling controlled camera motion without permanent modifications to the robotic end-effector. For each scene, a predefined and repeatable robot trajectory was executed. The acquisition procedure consisted of two stages: first, the robot grasped the stick-mounted camera; second, it followed a fixed scanning trajectory around the scene while recording synchronized RGB and depth images. This setup ensured consistent viewpoints across captures while maintaining sufficient viewpoint diversity for neural reconstruction methods.
        
        The poses of the cameras were initially estimated using COLMAP~\cite{schoenberger2016sfm}. However, COLMAP reconstructs scenes only up to an unknown similarity transformation, meaning that the global scale and absolute orientation are not directly recovered. To register each scene in a global metric coordinate frame, we placed four ArUco fiducial markers on the table surface. Markers were detected in captured images and used to estimate the reference frame of the scene, including its origin, orientation, and metric scale. Using the known physical dimensions and relative configuration of the markers, we resolved the scale ambiguity and aligned the reconstruction to a real-world coordinate system. This procedure allowed all reconstructed scenes to be expressed in meters, enabling direct geometric evaluation and quantitative comparison between different reconstruction methods.

    \subsection{Ground Truth Matching}
        For ground-truth geometry, we used high-precision laser scans provided by the YCB object data set~\cite{Calli2017ycb}. Each object in the data set is accompanied by a dense and metrically accurate mesh obtained through precise 3D scanning. These meshes serve as our geometric reference. Since our evaluation is performed on point clouds, we convert each ground-truth mesh into a point cloud by extracting all mesh vertices. Due to the high density of the original scans, the resulting point clouds provide a faithful approximation of the true object surfaces. To enable direct comparison with reconstructed scenes, ground-truth object meshes are transformed into the global metric coordinate system. For each scene, we align the meshes so that they correspond precisely to their real-world counterparts. 

    \subsection{Cloud Comparisons and Evaluation Metrics}

        To produce 3D models of the scene, we ran the selected methods with their default settings. For quantitative evaluation of geometric reconstruction quality, we restricted each point cloud to the bounding box containing the object. Additionally, we remove all points below the known table height. All point clouds are expressed in the same global metric coordinate frame prior to evaluation. We also denote $\mathcal{P} \subset \mathbb{R}^3$ as the reconstructed point cloud and $\mathcal{G} \subset \mathbb{R}^3$ as the point cloud of the ground-truth. We compare reconstructed point clouds against ground-truth laser scans in a common metric coordinate frame. All distances are reported in millimeters. We employ complementary metrics that measure surface accuracy, completeness, dispersion, and threshold-based correspondence. We assume that for each experiment the COLMAP introduces a similar error to the estimated one camera trajectory, thus equally influencing the compared reconstruction methods.
        
        \paragraph{Chamfer Distance (CD).}
        We evaluate geometric discrepancy using the (non-squared) Chamfer Distance between the reconstructed point cloud $\mathcal{P}$ and the ground-truth cloud $\mathcal{G}$. It consists of two directional components:
        \[
        \mathrm{CD}_{\mathcal{P}\rightarrow\mathcal{G}} =
        \frac{1}{|\mathcal{P}|} \sum_{p \in \mathcal{P}}
        \min_{g \in \mathcal{G}} \|p - g\|_2,
        \qquad
        \mathrm{CD}_{\mathcal{G}\rightarrow\mathcal{P}} =
        \frac{1}{|\mathcal{G}|} \sum_{g \in \mathcal{G}}
        \min_{p \in \mathcal{P}} \|g - p\|_2.
        \]
        The first term (reconstruction-to-ground-truth) measures surface accuracy, i.e., how precisely reconstructed points align with the reference geometry. The second term (ground-truth-to-reconstruction) measures completeness, indicating how well the reconstruction covers the true surface. 
        
        \paragraph{Chamfer Error Standard Deviation (Std@$\tau$).}
        To assess local geometric consistency, we compute the standard deviation of the reconstruction-to-ground-truth distances 
        $\min_{g \in \mathcal{G}} \|p-g\|_2$ 
        for points whose error is below a tolerance $\tau$ (2\,mm or 5\,mm). 
        This metric captures the spread of small-scale surface deviations and reflects reconstruction noise. 
        Lower values indicate more stable and locally consistent surfaces.
        
        \paragraph{Precision, Recall, and F1 Score.}
        For a distance threshold $\tau$, precision and recall are defined as:
        \[
        \text{Precision} =
        \frac{|\{ p \in \mathcal{P} \; | \; \min_{g \in \mathcal{G}} \|p-g\|_2 < \tau \}|}
        {|\mathcal{P}|},
        \]
        \[
        \text{Recall} =
        \frac{|\{ g \in \mathcal{G} \; | \; \min_{p \in \mathcal{P}} \|g-p\|_2 < \tau \}|}
        {|\mathcal{G}|}.
        \]
        Precision measures reconstruction correctness, while recall measures surface recovery. The F1 score is their harmonic mean:
        \[
        F1 = \frac{2 \cdot \text{Precision} \cdot \text{Recall}}
        {\text{Precision} + \text{Recall}}.
        \]
        We report results at 2\,mm and 5\,mm tolerances, reflecting precision requirements relevant to robotic manipulation.
        

\section{Results}


\subsection{Quantitative results}

\begin{table}[t]
\centering
\small
\begin{tabular}{l c c c c}
\toprule
\textbf{Method}
& $\mathrm{CD}_{\mathcal{P}\rightarrow\mathcal{G}} \downarrow$
& \textbf{Std@2mm} $\downarrow$
& \textbf{Std@5mm} $\downarrow$
& $\mathrm{CD}_{\mathcal{G}\rightarrow\mathcal{P}} \downarrow$ \\
\midrule
Rad-Meshes \cite{mai2026radiancemeshes} & \cc{0} 28.83 & \cc{5} 0.55 & \cc{30} 1.28 & \cc{0} 15.78 \\
Tri-Splats \cite{triangle_splatting} & \cc{5} 28.05 & \cc{10} 0.54 & \cc{0} 1.39 & \cc{100} \textbf{2.74} \\
COLMAP \cite{schoenberger2016sfm} & \cc{60} 13.02 & \cc{35} 0.53 & \cc{35} 1.25 & \cc{80} 5.16 \\
$\text{Rad-Meshes}_{pcd}$ \cite{mai2026radiancemeshes} & \cc{65} 11.72 & \cc{10} 0.54 & \cc{30} 1.28 & \cc{90} 3.79 \\
Inst-NGP \cite{instant-ngp} & \cc{85} 6.47 & \cc{25} 0.54 & \cc{25} 1.30 & \cc{90} 3.87 \\
3DGS \cite{Kerbl2023gaussian_splatting} & \cc{90} 5.24 & \cc{0} 0.55 & \cc{20} 1.31 & \cc{90} 4.05 \\
2DGS \cite{2d_gaussian} & \cc{100} \textbf{3.15} & \cc{5} 0.54 & \cc{30} 1.28 & \cc{85} 4.92 \\
Nerfacto \cite{nerfstudio_2023} & \cc{100} 3.78 & \cc{30} 0.53 & \cc{30} 1.27 & \cc{90} 3.91 \\
SVRaster \cite{sparse_voxels} & \cc{100} 3.18 & \cc{100} \textbf{0.50} & \cc{100} \textbf{1.00} & \cc{90} 4.32 \\
\bottomrule
\end{tabular}
\caption{
Surface reconstruction accuracy against laser scan ground truth.
}
\label{tab:accuracy}
\end{table}

\begin{table}[t]
\centering
\small
\begin{tabular}{l ccc ccc}
\toprule
\textbf{Method} & \multicolumn{3}{c}{@2mm $\uparrow$} & \multicolumn{3}{c}{@5mm $\uparrow$} \\
\cmidrule(lr){2-4} \cmidrule(lr){5-7}
& Prec & Rec & F1 & Prec & Rec & F1 \\
\midrule
Tri-Splats \cite{triangle_splatting} & \cc{0} 0.23 & \cc{100} \textbf{0.70} & \cc{45} 0.33 & \cc{0} 0.46 & \cc{100} \textbf{0.82} & \cc{45} 0.57 \\
Rad-Meshes \cite{mai2026radiancemeshes} & \cc{10} 0.28 & \cc{0} 0.10 & \cc{0} 0.14 & \cc{20} 0.54 & \cc{0} 0.30 & \cc{0} 0.36 \\
3DGS \cite{Kerbl2023gaussian_splatting} & \cc{40} 0.40 & \cc{45} 0.37 & \cc{50} 0.36 & \cc{65} 0.75 & \cc{85} 0.74 & \cc{80} 0.73 \\
$\text{Rad-Meshes}_{pcd}$ \cite{mai2026radiancemeshes} & \cc{40} 0.40 & \cc{60} 0.47 & \cc{60} 0.41 & \cc{65} 0.73 & \cc{90} 0.76 & \cc{85} 0.74 \\
Inst-NGP \cite{instant-ngp} & \cc{40} 0.40 & \cc{55} 0.44 & \cc{60} 0.40 & \cc{55} 0.70 & \cc{90} 0.78 & \cc{80} 0.72 \\
COLMAP \cite{schoenberger2016sfm} & \cc{45} 0.43 & \cc{20} 0.23 & \cc{35} 0.29 & \cc{50} 0.67 & \cc{65} 0.65 & \cc{65} 0.65 \\
2DGS \cite{2d_gaussian} & \cc{50} 0.46 & \cc{30} 0.28 & \cc{45} 0.33 & \cc{85} 0.82 & \cc{80} 0.71 & \cc{85} 0.74 \\
Nerfacto \cite{nerfstudio_2023} & \cc{60} 0.50 & \cc{65} 0.48 & \cc{75} 0.48 & \cc{85} 0.82 & \cc{90} 0.78 & \cc{95} 0.79 \\
SVRaster \cite{sparse_voxels} & \cc{100} \textbf{0.67} & \cc{70} 0.53 & \cc{100} \textbf{0.58} & \cc{100} \textbf{0.89} & \cc{85} 0.75 & \cc{100} \textbf{0.81} \\
\bottomrule
\end{tabular}
\caption{
Precision, recall and F1 at 2~mm and 5~mm tolerances. 
}
\label{tab:precision}
\vspace{-0.5cm}
\end{table}

The comparison focuses exclusively on the geometric accuracy. The results were obtained in 19 scenes, each containing 274 images with a resolution of 1280$\times$720\footnote{The data set with RGB images is available at~\url{https://doi.org/10.17632/zv9rk977wc.1}}. Table~\ref{tab:accuracy} summarizes surface accuracy and completeness metrics, while Table~\ref{tab:precision} reports threshold-based precision, recall, and F1 scores at tolerances of 2\,mm and 5\,mm.

\paragraph{Surface Accuracy.}
2DGS achieves the lowest $\mathrm{CD}_{\mathcal{P}\rightarrow\mathcal{G}}$ (3.15\,mm), closely followed by SVRaster (3.18\,mm) and Nerfacto (3.78\,mm). SVRaster obtains the lowest Chamfer error standard deviation, achieving the best Std@2\,mm (0.50) and Std@5\,mm (1.00), indicating more locally consistent reconstructions. Methods producing dense but unfiltered point clouds (Tri-Splats at 28.05\,mm, Rad-Meshes at 28.83\,mm) exhibit the highest mean surface deviation, while COLMAP (13.02\,mm) is limited by its sparse reconstruction.

\paragraph{Completeness and Coverage.}
Tri-Splats achieves the best $\mathrm{CD}_{\mathcal{G}\rightarrow\mathcal{P}}$ (2.74\,mm) and the highest Recall@2\,mm (0.70) and Recall@5\,mm (0.82), indicating that a lot of points lie near the surface. $\text{Rad-Meshes}_{pcd}$ (3.79\,mm), Inst-NGP (3.87\,mm), and Nerfacto (3.91\,mm) also demonstrate strong completeness. In contrast, the mesh-based Rad-Meshes variant shows significantly degraded completeness (15.78\,mm) and recall, suggesting geometric inconsistencies in its mesh extraction.

\paragraph{Threshold-Based Correspondence.}
At 2\,mm tolerance, SVRaster achieves the highest Precision@2\,mm (0.67) and F1@2\,mm (0.58), while Tri-Splats obtains the highest Recall@2\,mm (0.70), reflecting strong surface recovery but lower precision (0.23). Nerfacto provides a strong balance at this threshold with the second-highest F1@2\,mm (0.48). At 5\,mm, SVRaster achieves the best F1@5\,mm (0.81) and the highest Precision@5\,mm (0.89), followed by Nerfacto (F1@5\,mm 0.79, Precision@5\,mm 0.82), indicating a favorable balance between correctness and completeness at manipulation-relevant tolerances.

\paragraph{Computational Cost}

\begin{table}[t]
\centering
\caption{Average training time for the compared methods [s].}
\resizebox{\linewidth}{!}{
\begin{tabular}{c c c c c c c c}
\hline
Method & Rad-Meshes & Tri-Splats & 2DGS & Inst-NGP & 3DGS & Nerfacto & SVRaster \\
\hline
Time [s] & 18790.15 & 2788.82 & 1075.33 & 1046.35 & 885.71 & 644.30 & 568.72 \\
\hline
\end{tabular}
}
\label{tab:computation}\vspace{-0.75cm}
\end{table}

The computation cost analysis was performed on a computer with NVIDIA GeForce RTX 3090 and an Intel i9-9900KF CPU. The results in Tab.~\ref{tab:computation} show large differences in training time between methods. Rad-Meshes and Tri-Splats are the most computationally expensive, while Gaussian based methods significantly reduce training time. Nerfacto and SVRaster are the most efficient, with SVRaster achieving the lowest training time overall.

\subsection{Qualitative comparison}

\begin{figure}
    \centering
    \includegraphics[width=0.91\linewidth]{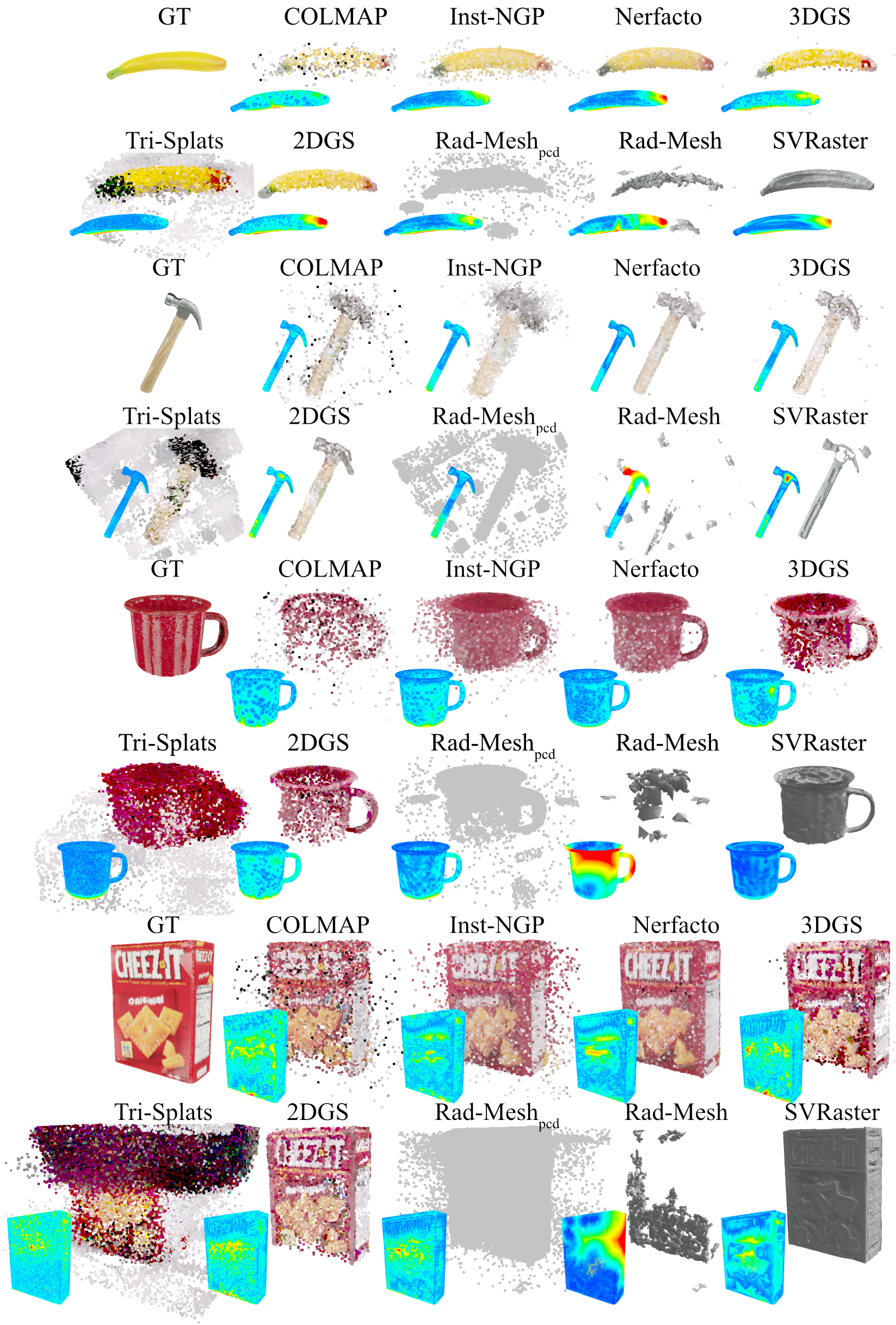}
    \caption{Qualitative comparison between point clouds obtained using considered scene models. Color map visualizes Chamfer error of point clouds.}
    \label{fig:qualitative}
\end{figure}

The qualitative comparison shown in Fig. 1 aligns with the quantitative evaluation results. Tri-Splats achieves the highest coverage. However, the resulting point cloud suffers from low accuracy. The point cloud generated by COLMAP is extremely sparse. In contrast, 2DGS and 3DGS produce denser point clouds, but their accuracy is limited. The most visually plausible 3D reconstructions are produced by SVRaster and Nerfacto.

\section{Conclusions}

The results reveal a clear trade off between geometric precision and surface completeness. Classical multi view reconstruction such as COLMAP, which relies on feature points extracted from RGB images, produces sparse point clouds and therefore provides limited surface coverage. Gaussian based approaches such as 3DGS and 2DGS generate denser point clouds with improved geometric accuracy, with 2DGS achieving one of the lowest surface deviations. However, the lack of an efficient Gaussian mixture sampling strategy leads to gaps in the reconstructed models. The Tri-Splats method often deviates from the true geometry and fails during reconstruction, resulting in noisy and unusable point clouds.

In contrast, neural radiance field approaches such as Nerfacto, together with SVRaster, produce visually plausible reconstructions and offer a better balance between geometric accuracy and completeness. Nerfacto integrates techniques developed in NeRF based research, including improved density field sampling, camera pose refinement, and piecewise sampling strategies, and achieves the best performance among the evaluated NeRF based methods. SVRaster, on the other hand, adaptively fits sparse voxels to different levels of scene detail. Its octree based structure reduces memory consumption while preserving fine geometric details. Our experiments show that SVRaster and Nerfacto are preferable for robotic tasks that rely on accurate 3D scene representations.

\section*{Acknowledgment}
\scriptsize{The work was supported by the National Science Centre, Poland, under research project no UMO-2023/51/B/ST6/01646. The Computations were performed using the infrastructure of the Poznan Supercomputing and Networking Center (PCSS).}

\vspace{-0.15cm}
\bibliography{pprai}
\bibliographystyle{plain}

\end{document}